\documentclass{article}
\usepackage{ijcai09}

\usepackage{times}

\usepackage{latexsym} 

\usepackage{graphicx}

\newtheorem{mytheorem}{Theorem}
\newcommand{\myproof}{\noindent {\bf Proof:\ \ }}

\newcommand{\myqed}{\mbox{$\Box$}}

\newcommand{\myOmit}[1]{}

\title{Where are the really hard manipulation problems?\\
The phase transition in manipulating the veto rule\thanks{NICTA is 
funded by the Australian Government through the Department of Broadband, Communications and the Digital Economy and the Australian Research Council through the ICT Centre of Excellence program.}}
\author{Toby Walsh\\
NICTA and UNSW\\
Sydney, Australia\\
toby.walsh@nicta.com.au}

\begin{document}

\maketitle

\begin{abstract}
Voting is a simple mechanism to aggregate the preferences of
agents. Many voting rules have been shown to be NP-hard
to manipulate. However, a number of recent 
theoretical results suggest that 
this complexity may only be in the worst-case 
since manipulation is often easy in practice. In this paper,
we show that empirical studies are useful in 
improving our understanding of this issue. 
We demonstrate that there is a smooth transition in the probability
that a coalition can 
elect a desired candidate using the veto rule as the size of 
the manipulating coalition increases. We show
that a rescaled probability curve displays a simple and universal form 
independent of the size of the problem. We argue that 
manipulation of the veto rule is asymptotically
easy for many independent and identically distributed votes
even when the
coalition of manipulators is critical in size. 
Based on this argument, we identify a situation
in which manipulation is computationally hard. 
This is when votes are highly correlated
and the election is ``hung''. 
We show, however, that
even a single uncorrelated voter is enough to 
make manipulation easy again. 
\end{abstract}

\section{Introduction}

The Gibbard-Satterthwaite theorem proves that, under some
simple assumptions, most voting rules
are manipulable. That is, it may pay for an agent not to
report their preferences truthfully. 
One possible escape from 
this result was proposed by Bartholdi, Tovey and Trick 
\cite{bartholditoveytrick}.
Whilst a manipulation may exist, perhaps it is computationally
too difficult to find. Many results have subsequently
been proven showing that various voting rules are NP-hard 
to manipulate under different assumptions
including: an unbounded number of candidates; a small number of candidates
but weighted votes; and uncertainty in the distribution
of votes. See, for instance, \cite{bartholditoveytrick,stvhard,csljacm07}.
There is, however, increasing concern that worst-case results
like these may not reflect the difficulty of manipulation
in practice. Indeed, a number of recent theoretical results 
suggest that manipulation may often be computationally
easy \cite{csaaai2006,prjair07,xcec08,fknfocs09,xcec08b}. 

In this paper we show that, in addition to
attacking this question theoretically, we can
profitably study it empirically. There are several reasons why empirical
analysis is useful. First, theoretical 
analysis is often asymptotic so does not
show the size of hidden constants. 
In addition, elections are typically bounded in
size. Can we be sure that asymptotic behaviour
is relevant for the finite sized electorates met in practice? 
Second, theoretical analysis is often
restricted to particular distributions (e.g. independent
and identically distributed votes). 
Manipulation may be very different in practice due to 
correlations between votes. For instance, 
if all preferences are single-peaked then there
are voting rules which cannot be manipulated. 
It is in the best interests of all agents to state
their true preferences. 
Third, many of these theoretical results about the
easiness of manipulation have been hard won 
and are limited in their scope. 
For instance, Friedgut et al. were not able to extend their
result beyond three candidates \cite{fknfocs09}.
An empirical study may quickly suggest if the result
extends to more candidates. 
Finally, empirical studies may suggest new avenues
for theoretical study. For example, the experiments
reported here suggest a simple and universal
form for the probability that a coalition
is able to elect a desired candidate. 
It would be interesting to try to derive 
this form theoretically. 

\section{Background}

We suppose that there are $n$ agents who have voted
and a coalition of $m$ additional agents who 
wish to manipulate the result. 
When the manipulating coalition is small, they have too little 
weight to be able to change the result. On the other
hand, when the coalition is large, they are sure
to be able to make their desired candidate win.
Procaccia and Rosenschein proved that for most
scoring rules and a wide variety of
distributions over votes, 
when $m = o(\sqrt{n})$, the probability
that a manipulating coalition can change the result 
tends to 0, and when $m = \omega(\sqrt{n})$, the probability
that they can manipulate the result
tends to 1 \cite{praamas2007}.
They offer two interpretations of this result. On the positive
side, they suggest it may focus attention on other distributions
which are computationally hard to manipulate. 
On the negative side, they suggest that it may strengthen the 
argument that manipulation problems are easy on
average. 

More recently, Xia and Conitzer have shown that for a large class
of voting rules, as the number of agents grows, either 
the probability that a coalition
can manipulate the result is very small (as the
coalition is too small), or the
probability that they can easily manipulate the result
to make any alternative
win is very large \cite{xcec08}. 
They leave open only a small interval in the size
of the coalition for which the coalition is 
large enough to be able to manipulate but
not obviously large enough to be able to manipulate the
result easily. 
More precisely, for a wide range of voting
rules including scoring rules, STV, Copeland and
maximin, with votes which are drawn independently and with an
identical distribution that is positive
everywhere, they identify three cases:
\begin{itemize}
\item if $m=O(n^p)$ for $p<\frac{1}{2}$
then the probability that the result can be 
changed is $O(\frac{1}{\sqrt{n}})$;
\item if $m=\Omega(n^p)$ for $p>\frac{1}{2}$
and $o(n)$ and votes are uniform
then the probability that the result can be 
manipulated is $1-O(e^{-\Theta(n^{2p-1})})$ using
a simple greedy procedure;
\item if $m=\Theta(\sqrt{n})$
then they provide no result.
\end{itemize}
In this paper, we shall provide empirical evidence
to help close this gap and understand what happens
when the coalition is of a critical size that
grows as $\Theta(\sqrt{n})$. 

\section{Finding manipulations}

We focus on the veto rule. This is a scoring rule
in which each agent gets to cast a veto against one candidate. 
The candidate with the fewest vetoes wins. 
We suppose that tie-breaking is in favor of the manipulators.
However, it is easy to relax this assumption. 
There are several
reason why we start this
investigation into the complexity of manipulation 
with the veto rule. First, the veto rule is very
simple to reason about. This can be contrasted
with other voting rules that are computationally
hard to manipulate. For example, the STV rule
is NP-hard to manipulate \cite{stvhard} 
but this complexity appears to come from reasoning about 
what happens between the different rounds. 
Second, the veto rule is on the borderline of tractability
since constructive manipulation of the rule by a coalition of
weighted agents is NP-hard but destructive manipulation
is polynomial \cite{csljacm07}. 
Third, as the next theorem shows, number
partitioning algorithms can be used to compute
a successful manipulation of the veto rule. 
More precisely, manipulation of an election
with 3 candidates and weighted votes (which is 
NP-hard \cite{csljacm07}) can be directly reduced
to 2-way number partitioning. We therefore 
compute manipulations in our
experiments using the efficient CKK algorithm 
\cite{korf2}. 

\begin{mytheorem}
There exists a successful manipulation
of an election with 3 candidates by a weighted coalition 
using the veto rule iff 
there exists a partitioning
of $W \cup \{|a-b|\}$ into two bags
such that the difference between their
two sums is less than or equal to 
$a+b-2c + \sum_{i \in W} i$
where $W$ is the multiset of weights of the 
manipulating coalition, $a$, $b$ and $c$
are the weights of vetoes assigned to
the three candidates by the non-manipulators
and the manipulators wish the candidate with weight $c$
to win. 
\end{mytheorem}
\myproof
It never helps a coalition manipulating the veto rule
to veto the candidate that they wish to win. 
The coalition does, however, need to decide how
to divide their vetoes between the candidates that they wish
to lose. Consider the case $a \geq b$. 
Suppose the partition has
weights $w-\Delta/2$ and $w+\Delta/2$
where $2w=\sum_{i \in W \cup \{|a-b|\}} i$
and $\Delta$ is the difference between 
the two sums. 
The same partition of vetoes is a successful
manipulation iff the winning candidate has no more
vetoes than the next best candidate. That is,
$c \leq b+ (w-\Delta/2)$. 
Hence $\Delta \leq 2w+2b-2c = (a-b)+ 2b -2c + \sum_{i \in W} i = 
(a+b-2c) + 2\sum_{i \in W} i$. 
In the other case, $a < b$ and
$\Delta \leq (b+a-2c) + \sum_{i \in W} i$. 
Thus $\Delta \leq a+b-2c + \sum_{i \in W} i$. 
\myqed

Similar arguments can be given
to show that the manipulation of a veto election of $p$ candidates
can be reduced to 
finding a $p-1$-way partition of numbers,
and that manipulation of {\em any} scoring rule
with 3 candidates and weighted votes can be 
reduced to 2-way number partitioning.
However, manipulating elections with greater than 3 candidates
and scoring rules other than 
veto or plurality appears
to require other computational
approaches. 

\section{Uniform votes}

\begin{figure}[htb]
\vspace{-1in}
\begin{center}
\includegraphics{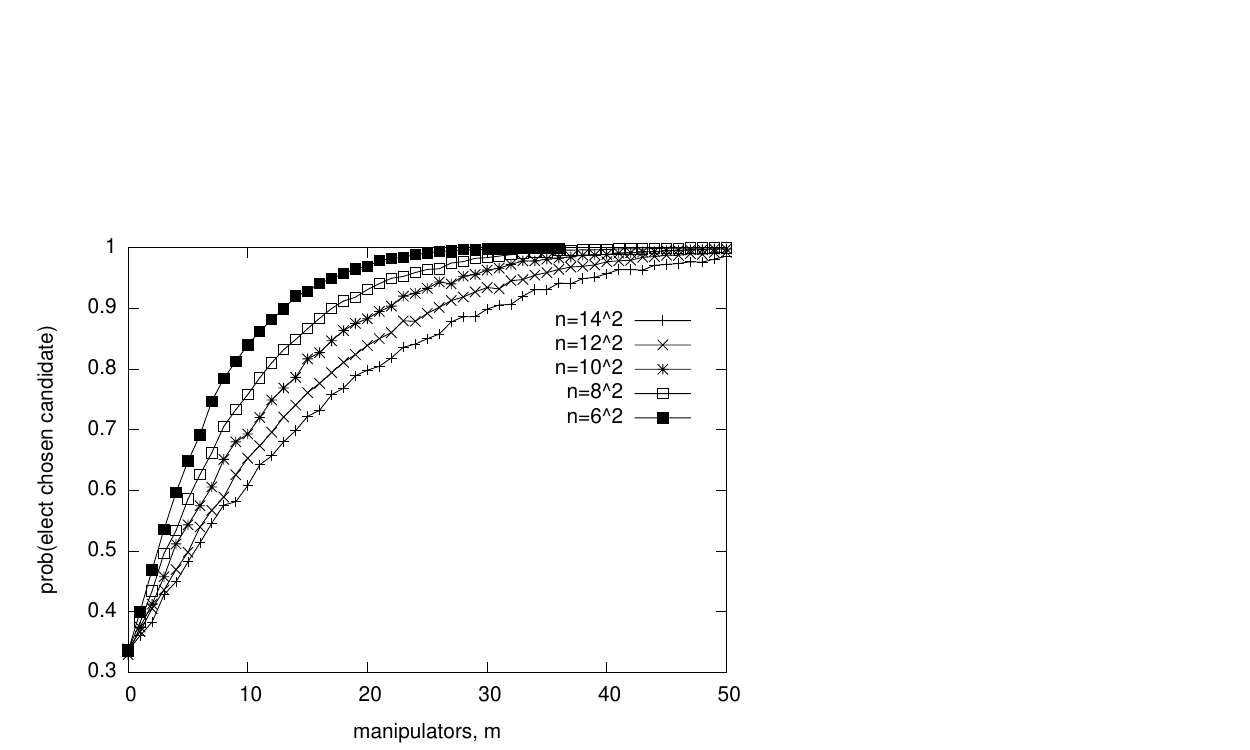}
\end{center}
\caption{Probability of 
a coalition of $m$ agents electing a chosen candidate
where $n$ agents have already voted.
Vetoes are weighted and uniformly
drawn from $(0,2^8]$. At $m=0$, there
is a 1/3rd chance that the non-manipulators
have already elected this candidate. 
In this and all subsequent experiments,
we tested 10,000 problems at each data point.}
\label{fig-prob}
\end{figure}

We consider the case that 
the $n$ agents veto uniformly at random one of the
3 possible candidates, 
and vetoes carry weights drawn uniformly from $(0,k]$. 
When the coalition is small in size, it has too little 
weight to be able to change the result. On the other
hand, when the coalition is large in size, it is sure
to be able to make a favored candidate win.
There is thus a transition in the manipulability
of the problem as the coalition size increases
(see Figure \ref{fig-prob}). 

\begin{figure}[htb]
\vspace{-1in}
\begin{center}
\includegraphics{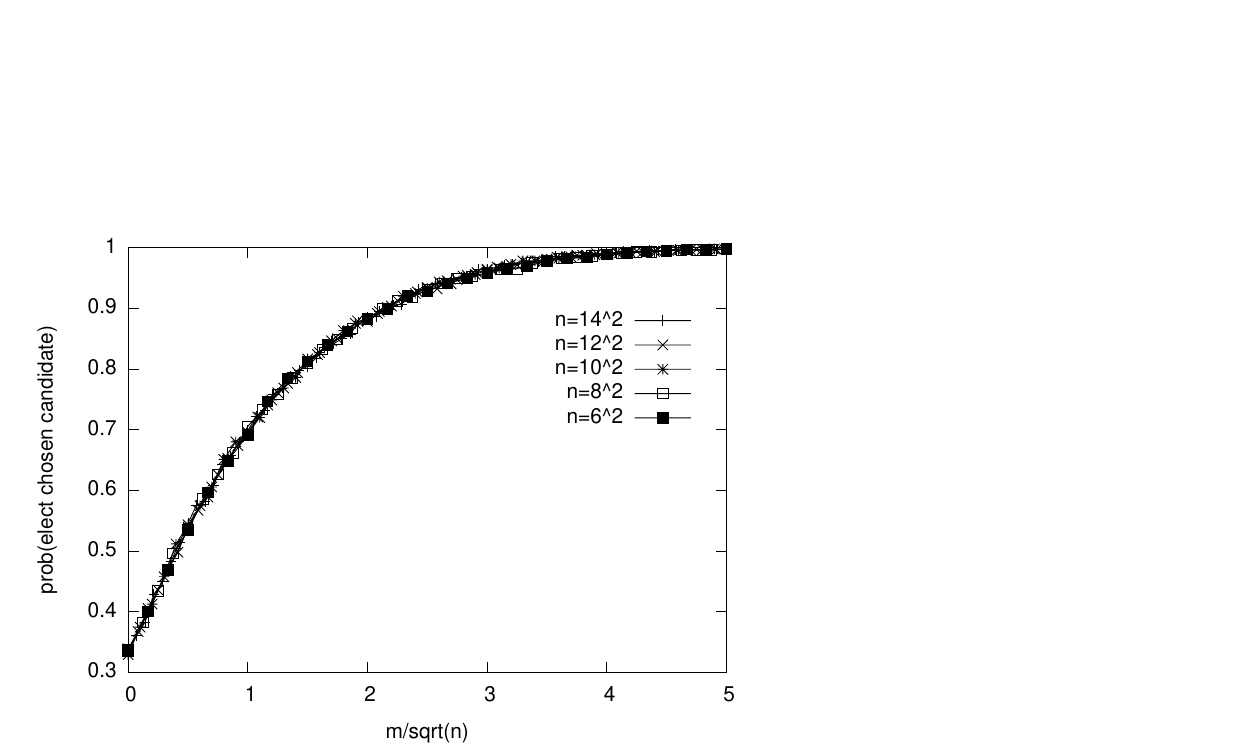}
\caption{Rescaled probability that a coalition of $m$ agents
can elected a chosen candidate where $n$ agents
have already voted. Vetoes are weighted and uniformly
drawn from $(0,2^8]$. The x-axis is scaled by $1/\sqrt{n}$. }
\label{fig-rescale}
\end{center}
\end{figure}
Based on 
\cite{praamas2007,xcec08}, we expect the critical coalition
size to increase as $\sqrt{n}$. 
In Figure \ref{fig-rescale}, we see
that the phase transition 
displays a simple and universal form
when plotted against $m/\sqrt{n}$. 
The phase transition appears to be smooth,
with the probability varying slowly and 
not approaching a step function as problem
size increases. 
We obtained a good fit with $1- \frac{2}{3}e^{-m/\sqrt{n}}$. 
Other smooth phase transitions have
been seen with 2-coloring \cite{achlioptasphd}, 1-in-2 satisfiability
and Not-All-Equal 2-satisfiability \cite{waaai2002}.
It is interesting to note that all these
decision problems are polynomial.

\begin{figure}[htb]
\vspace{-1in}
\begin{center}
\includegraphics{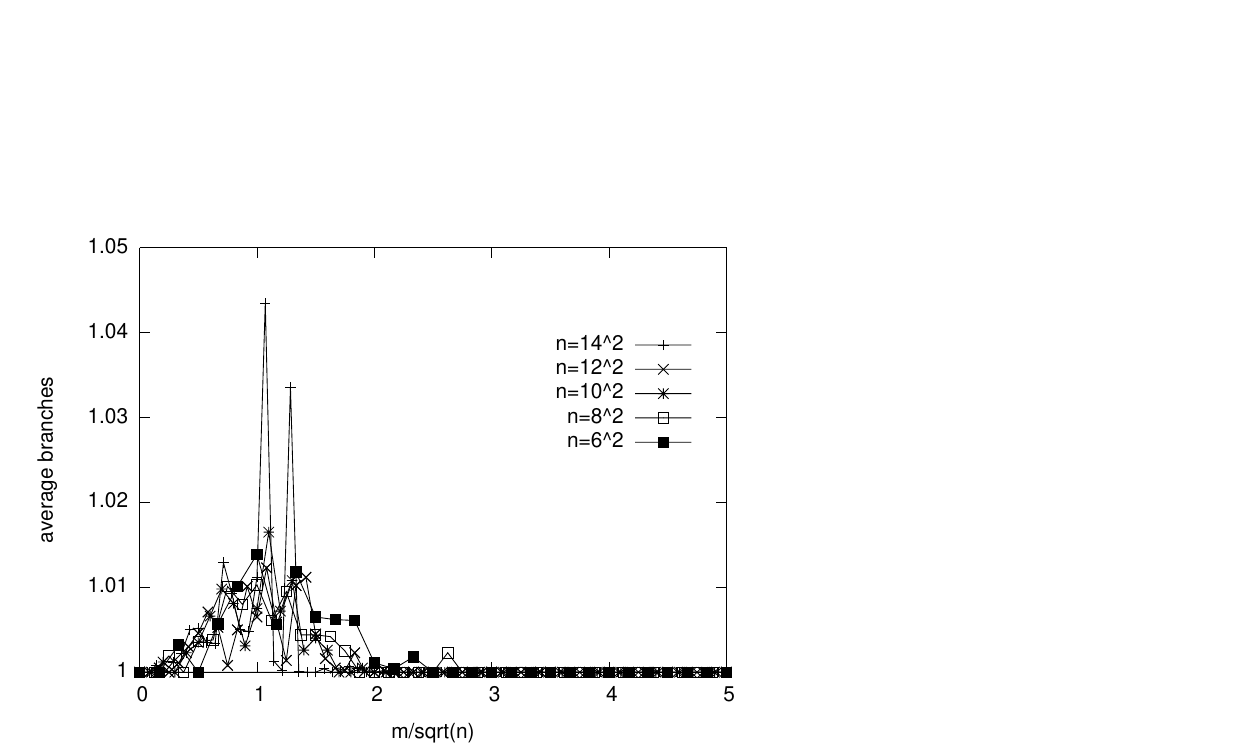}
\caption{Computational cost for the CKK algorithm
to decide if a coalition of $m$ agents
can manipulate a veto election where $n$ agents
have already voted. Vetoes are weighted and uniformly
drawn from $(0,2^m]$. All problems are
solved with little search. 
}
\label{fig-branches}
\end{center}
\end{figure}

The theoretical results mentioned earlier leave
open how hard it is to compute whether a manipulation
is possible 
when the coalition size is critical. 
Figure \ref{fig-branches} displays the computational
cost to find a manipulation (or prove none exists) using
the efficient CKK algorithm. 
Even in the critical region
where problems may or may not be manipulable, it is 
easy to compute whether the problem is manipulable. 
All problems are solved in a few branches. 
This contrasts with phase transition behaviour in problems like
satisfiability 
\cite{cheeseman-hard,mitchell-hard-easy,gw-ecai94},
constraint satisfaction \cite{csp-pt},
number partitioning \cite{gw-ecai96,gw-ci98}
and the traveling salesman problem \cite{gw-tsp96}
where the hardest problems 
occur around the phase transition. 

\section{Why hard problems are rare}

Based on our reduction of manipulation 
problems to number partitioning, we 
give a heuristic argument 
why hard manipulation problems become vanishing
rare as $n \leadsto \infty$ and $m = \Theta(\sqrt{n})$. 
The basic idea is simple: by the time the coalition
is large enough to be able to change the result,
the variance in scores between the candidates
is likely to be so large that computing a 
successful manipulation or proving none is possible will be easy. 

Suppose that the manipulators
want candidates $A$ and $B$ 
to lose so that $C$ wins, and 
that the non-manipulators have cast vetoes of
weight $a$, $b$ and $c$ for 
$A$, $B$ and $C$ respectively. 
Without loss of generality we suppose 
that $a \geq b$. 
There are three cases to consider.
In the first case, $a \geq c$ and $b \geq c$. 
It is then easy for the manipulators to make $C$
win since $C$ wins whether
they veto $A$ or $B$. In the second case, 
$a \geq c > b$. 
Again, it is easy for the manipulators to 
decide if they can make $C$
win. They all veto $B$. There is a successful
manipulation iff $C$ now wins. 
In the third  case, $a < c$ and $b < c$. 
The manipulators must partition
their $m$ vetoes between $A$ and $B$ so
that the total vetoes received by
$A$ and $B$ exceeds those for $C$. 
Let $d$ be the deficit
in weight between $A$ and $C$ and between $B$ and $C$. 
That is, $d=(c - a) + (c - b)=2c-a-b$.
We can model $d$ as the sum of $n$ 
random variables drawn uniformly with probability 
1/3 from $[0,2k]$ and with probability 2/3 from $[-k,0]$. 
These variables have mean 0 and variance ${2k^2}/{3}$. 
By the Central Limit
Theorem, $d$ tends to a normal distribution with 
mean 0, and variance $s^2  = {2nk^2}/{3}$.
For a manipulation to be possible, $d$ must 
be less than $w$, the sum of the weights of the
vetoes of the manipulators. 
By the Central Limit
Theorem, $w$ also tends to a normal distribution with 
mean $\mu = mk/2$, and variance $\sigma^2={2mk^2}/{3}$.

A simple heuristic argument due to \cite{karmarkar2}
and also based on the Central Limit Theorem
upper bounds the optimal partition difference 
of $m$ numbers from $(0,k]$ by $O(k \sqrt{m}/2^m)$. 
In addition, based on the phase transition in number partitioning \cite{gw-ci98},
we expect partitioning problems to be easy unless $\log_2(k) = \Theta(m)$. 
Combining these two observations, we expect hard manipulation
problems when
$0 \leq w-d \leq \alpha \sqrt{m}$ for some constant $\alpha$.
The probability of this occurring is:
$$ \int_0^\infty 
 \frac{1}{\sqrt{2\pi} \sigma} e^{-\frac{(x-\mu)^2}{2\sigma^2}} 
\int_{x-\alpha \sqrt{m}}^x \frac{1}{\sqrt{2\pi} s} e^{-\frac{y^2}{2s^2}} \ dy \ dx $$
By substituting for $s$, $\mu$ 
and $\sigma$, we get: 
$$ \int_0^\infty 
 \frac{1 }{\sqrt{4\pi mk^2/3}} e^{-\frac{(x-mk/2)^2}{4mk^2/3}} 
\int_{x-\alpha \sqrt{m}}^x \frac{1 }{\sqrt{4\pi nk^2/3} }  
e^{-\frac{y^2}{4nk^2/3}} \ dy \ dx 
$$
For $n \leadsto \infty$, this tends to:
$$ \int_0^\infty 
 \frac{1}{\sqrt{4\pi mk^2/3}} e^{-\frac{(x-mk/2)^2}{4mk^2/3}} 
\frac{\alpha \sqrt{m}}{\sqrt{4\pi nk^2/3} } e^{-\frac{x^2}{4nk^2/3}}  \ dx 
$$
As $e^{-z} \leq 1$ for $z>0$, this is upper bounded by:
$$ \frac{\alpha \sqrt{m}}{\sqrt{4\pi nk^2/3} } \int_0^\infty 
 \frac{1}{\sqrt{4\pi mk^2/3}} e^{-\frac{(x-mk/2)^2}{4mk^2/3}} 
\ dx 
$$
Since the integral is bounded by 1, 
$m = \Theta(\sqrt{n})$ and $\log_2(k) = \Theta(m)$, 
this upper bound varies as: 
$$ O(\frac{1}{\sqrt{m} 2^m})
$$
Thus, we expect hard instances of manipulation problems
to be exponentially rare. Since even a brute force
manipulation algorithm takes $O(2^m)$ time in the worst-case, 
we do not expect the hard instances to have
a significant impact on the average-case as $n$ 
(and thus $m$) grows. 
We stress this is only a heuristic argument.
It makes assumptions about the
complexity of manipulation problems (in particular
that hard instances should lie within the narrow 
interval $0 \leq w - d \leq \alpha \sqrt{m}$). These assumptions
are currently only supported by empirical observation and 
informal argument. However, the experimental results reported
in Figure \ref{fig-branches} support these conclusions. 

\section{Varying weights}

\begin{figure}[htb]
\vspace{-1in}
\begin{center}
\includegraphics{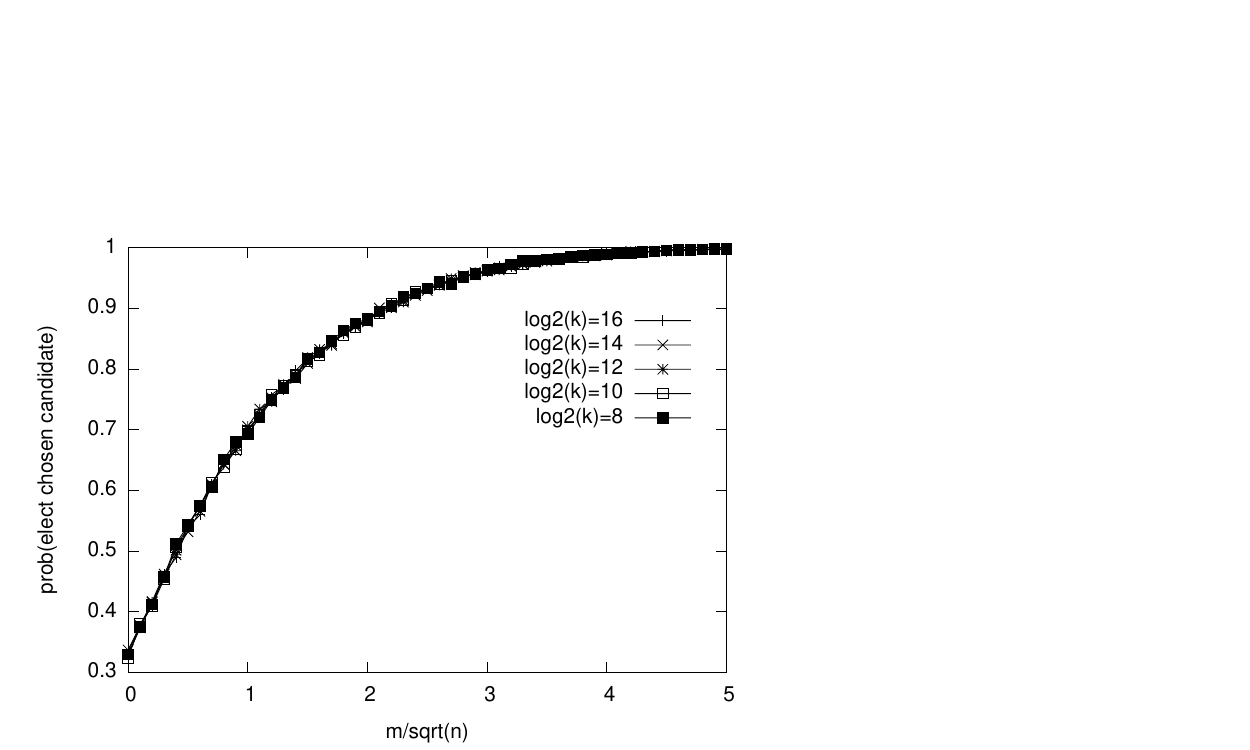}
\caption{Independence of the size of the 
weights and the manipulability of
an election. Probability that a coalition of $m$ agents
can elect a chosen candidate where $n$ agents
have already voted. Vetoes are weighted and uniformly
drawn from $(0,k]$. }
\label{fig-rescale-weights}
\end{center}
\end{figure}

The theoretical analyses of manipulation 
in \cite{praamas2007,xcec08}
suggest that the probability of an election being manipulable
is largely independent of $k$, the size of
the weights attached to the vetoes. 
Figure \ref{fig-rescale-weights} demonstrates
that this indeed appears to be the case in practice.
When weights are varied in size from $2^8$ to $2^{16}$,
the probability does not appear to change. 
In fact, the probability curve fits the same simple and
universal form plotted in Figure \ref{fig-rescale}. 
We also observed that the cost of computing
a manipulation or proving that none is possible
did not change as 
the weights were varied in size.

\section{Normally distributed votes}

What happens with other distributions of 
votes? 
The theoretical analyses of manipulation 
in \cite{praamas2007,xcec08}
suggest that there is a critical coalition
size that increases as $\Theta(\sqrt{n})$
for many types of independent and identically distributed random votes.
Similarly, our heuristic argument about 
why hard manipulation problems are vanishingly
rare depends on application of the Central Limit
Theorem. It therefore works with other types of 
independent and identically distributed random votes.

\begin{figure}[htb]
\vspace{-1in}
\begin{center}
\includegraphics{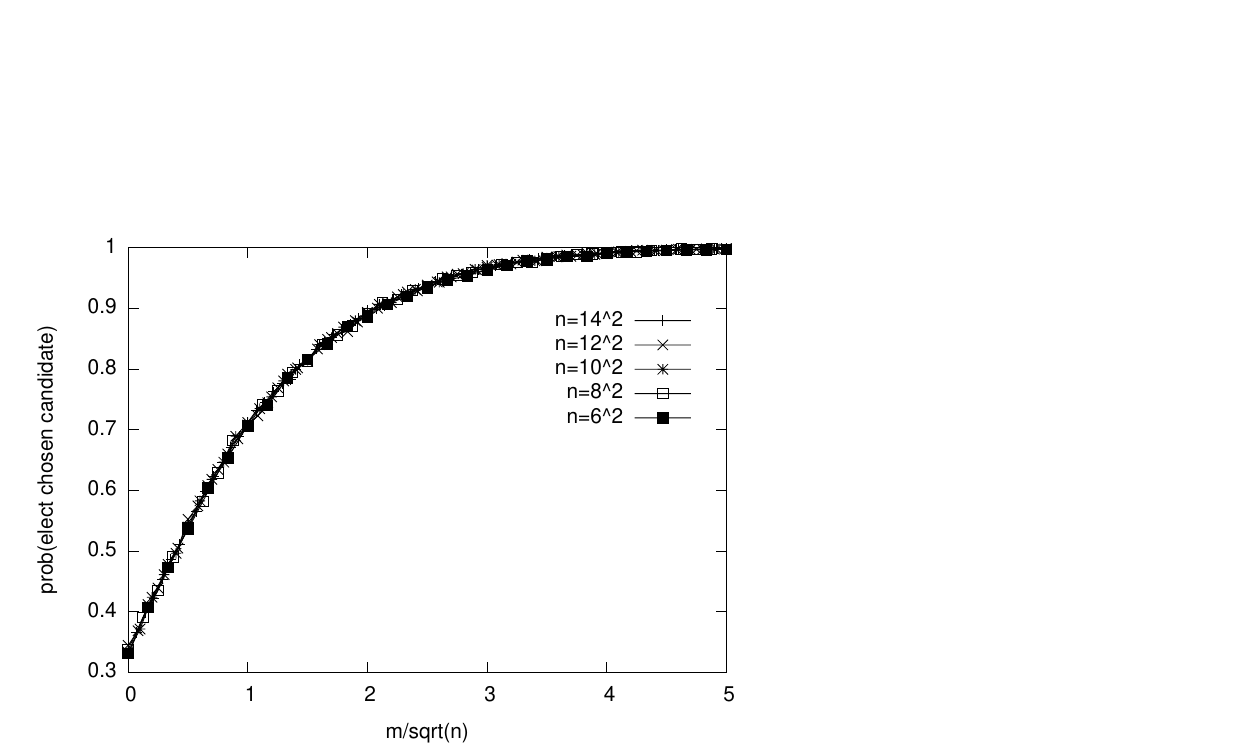}
\caption{Weighted votes taken from a normal 
distribution. We plot the probability that a coalition of $m$ agents
can elect a chosen candidate where $n$ agents
have already voted. Vetoes are weighted and 
drawn from a normal distribution
with mean $2^8$ and standard deviation $2^7$. 
The x-axis is scaled by $\sqrt{n}$. }
\label{fig-normal}
\end{center}
\end{figure}

We shall consider therefore another 
type of independent and identically distributed vote. 
In particular, we study an election in which
weights are independently drawn from a normal distribution. 
Figure \ref{fig-normal}
shows that there is again a smooth 
phase transition in manipulability.
We also plotted Figure \ref{fig-normal} on
top of Figures \ref{fig-rescale} and 
\ref{fig-rescale-weights}. All
curves appear to fit the same simple and universal form. 
As with uniform weights, the computational cost of deciding
if an election is manipulable was small
even when the coalition size was critical. 
Finally, we varied the parameters of the normal
distribution. The probability of electing a chosen candidate
as well as the cost of computing
a manipulation did not appear to depend on the mean or
variance of the distribution. 

\section{Correlated votes}

We conjecture that
one place to find hard manipulation problems is where
votes are more correlated.
\begin{figure}[htb]
\vspace{-1in}
\begin{center}
\includegraphics{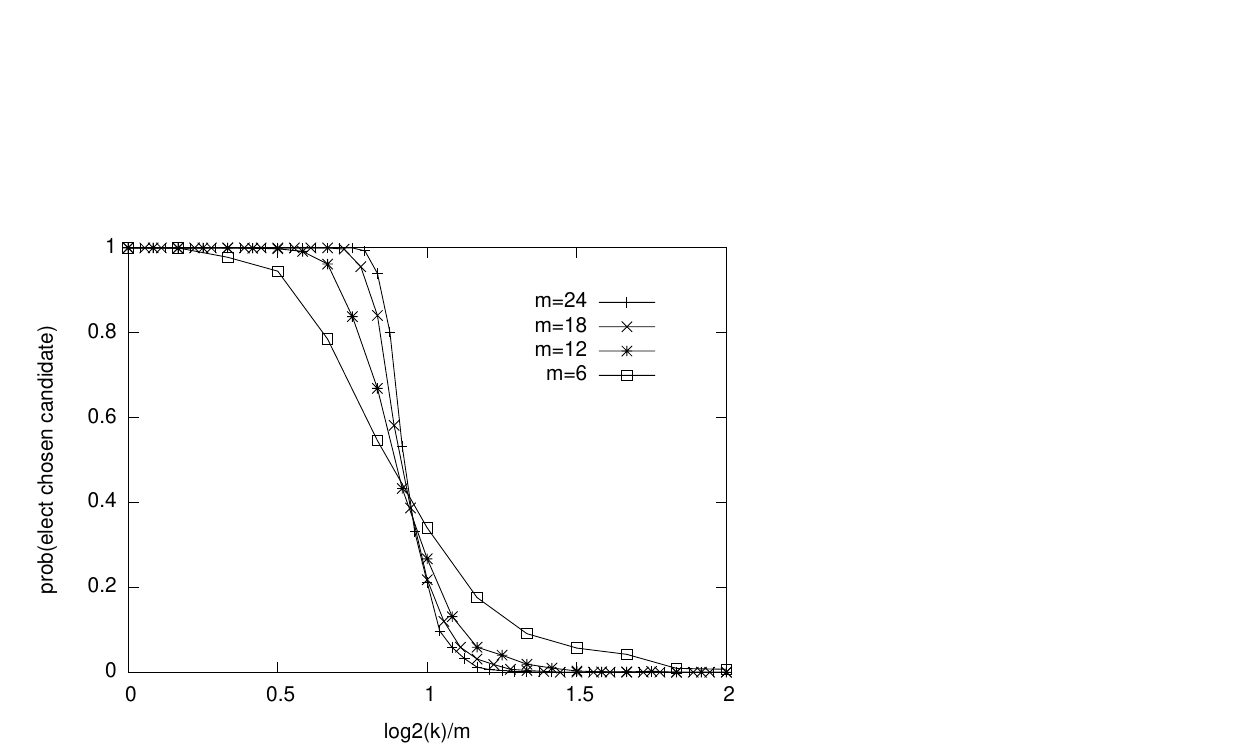}
\caption{Manipulation of an election where votes are
highly correlated and the result is ``hung''. 
We plot the probability that a coalition of $m$ agents
can elect a chosen candidate. 
Vetoes of the manipulators are weighted and uniformly
drawn from $(0,k]$, the other agents have all vetoed
the candidate that the manipulators wish to win,
and the sum of the weights of the manipulators is twice
that of the non-manipulators. }
\label{fig-hung}
\end{center}
\end{figure}
\begin{figure}[htb]
\vspace{-1in}
\begin{center}
\includegraphics{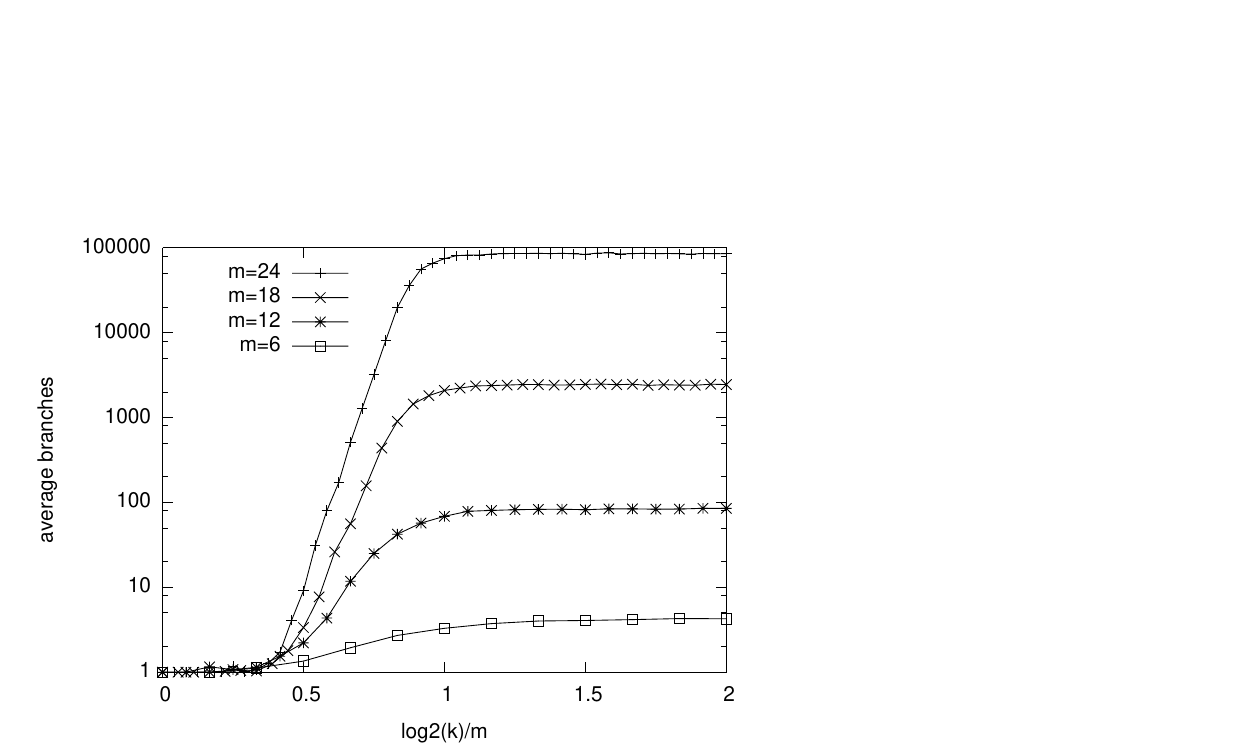}
\caption{The cost to decide if a hung election can be manipulated. 
We plot the cost for the CKK algorithm
to decide if a coalition of $m$ agents
can manipulate a veto election. 
Vetoes of the manipulators are weighted and uniformly
drawn from $(0,k]$, the other agents have all vetoed
the candidate that the manipulators wish to win,
and the sum of the weights of the manipulators is twice
that of the non-manipulators. }
\label{fig-hung-br}
\end{center}
\end{figure}
For example, consider a ``hung'' election where 
all $n$ agents veto the candidate that the manipulators
wish to win, but the $m$ manipulators have exactly
twice the weight of vetoes of the $n$ agents. 
This election is finely balanced. 
The favored candidate of the manipulators wins iff the manipulators
perfectly partition their vetoes between the two
candidates that they wish to lose. 
In Figure \ref{fig-hung},
we plot the probability that the $m$ manipulators can 
make their preferred candidate win in such
a ``hung'' election as we vary
the size of their weights $k$. Similar to 
number partitioning \cite{gw-ci98}, we see a rapid
transition in manipulability around $\log_2(k)/m \approx 1$. 
In Figure \ref{fig-hung-br},
we observe that there is a rapid increase
in the computationally complexity to compute
a manipulation around this point. 

\begin{figure}[htb]
\vspace{-1in}
\begin{center}
\includegraphics{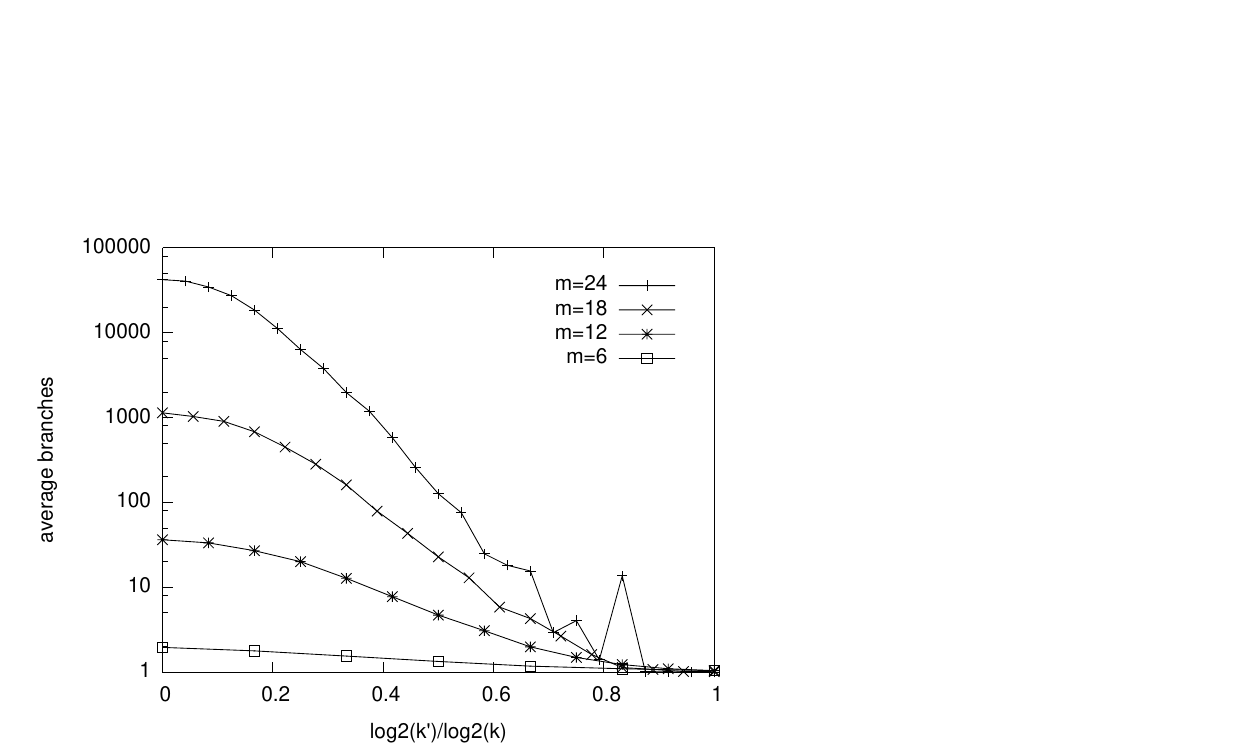}
\caption{The impact of one random voter on the manipulability
of a hung election. We plot the cost for the CKK algorithm
to decide if a coalition of $m$ agents
can manipulate a veto election.
Vetoes of the manipulators are weighted and uniformly
drawn from $(0,k]$, the non-manipulating agents have all vetoed
the candidate that the manipulators wish to win,
and the sum of the weights of the manipulators is twice
that of the non-manipulators except for one 
random non-manipulating agent
whose weight is uniformly drawn from $(0,k']$. 
When the veto of the one random voter has the same weight
as the other voters, it is computationally easy to 
decide if the election can be manipulated. }
\label{fig-1rand}
\end{center}
\end{figure}

What happens when the votes are less
correlated? We consider an election which is perfectly hung as before 
except for one agent who votes at random between
the three candidates. In Figure 
\ref{fig-1rand}, we plot the cost of
computing a manipulation as 
the weight of this single random veto increases. 
Even one uncorrelated vote is enough to
make manipulation easy if it has the same magnitude
in weight as the vetoes of the manipulators. This suggests
that we will only find hard manipulation
problems in when votes are highly correlated.

\section{Other related work}

There have been a number of other recent theoretical
results about the computational complexity of 
manipulating elections. For instance, 
Procaccia and Rosenschein give a simple
greedy procedure that will find a manipulation 
of a scoring rule for any ``junta'' distribution
of weighted votes in polynomial
time with a probability of failure that is 
an inverse polynomial in $n$
\cite{prjair07}. A ``junta'' distribution
is concentrated on the hard instances.

As a second example, 
Friedgut, Kalai and Nisan prove that
if the voting rule is neutral and
far from dictatorial and there
are 3 candidates then there exists
an agent for whom a random manipulation 
succeeds with probability $\Omega(\frac{1}{n})$
\cite{fknfocs09}. 
Xia and Conitzer showed that, starting
from different assumptions, a random
manipulation would succeed with probability
$\Omega(\frac{1}{n})$ for 3 or more 
candidates for STV, for 4 or more candidates for any scoring
rule and for 5 or more candidates for Copeland \cite{xcec08b}. 

Coleman and Teague provide polynomial
algorithms to compute a manipulation for
the STV rule when either the number of voters
or the number of candidates is fixed \cite{ctcats2007}.
They also conducted an empirical study
which demonstrates that only relatively small
coalitions are needed to change the elimination
order of the STV rule. They observe that most uniform
and random elections
are not trivially manipulable using a simple greedy
heuristic.

Finally, similar phenomena have been observed
in the phase transition
for the Hamiltonian cycle problem \cite{fgwipl,vcjair98}. 
If the number of edges is small, there is 
likely to be a node of degree smaller than 2.
There cannot therefore be any Hamiltonian cycle.
By the time that there are enough edges for all
nodes to be of degree 2, there are likely to be
many possible Hamiltonian cycles and even a simple
heuristic can find one. Thus, the phase transition
in the existence of a Hamiltonian cycle is not
associated with hard instances of the problem.
The behavior seen here is similar. 
By the time the coalition is large
enough to manipulate the result, the 
variance in scores between the candidates
is likely to be so large that computing a 
successful manipulation or proving none is possible is easy. 

\section{Conclusions}

We have studied
whether computational
complexity is a barrier to the manipulation
for the veto rule. We showed that there is a 
{\em smooth} transition in the probability
that a coalition can elect a desired candidate as the size of 
the manipulating coalition is varied. 
We demonstrated that
a rescaled probability curve displays a simple universal form 
independent of problem size.
Unlike phase transitions for other NP-complete problems,  
hard problems are not
associated with this transition. 
Finally, we studied the impact of correlation between
votes. We showed that manipulation is hard
when votes are highly correlated and the election is ``hung''. 
However, even one uncorrelated voter was enough to make 
manipulation easy again. 

What lessons can be learnt from this study?
First, there appears to be an universal
form for the probability that a coalition can 
manipulate the result. Can we
derive this theoretically? 
Second, whilst we have focused on the veto rule,
similar behavior is likely
with other voting rules. It would, for instance, be
interesting to study a more complex rule like STV 
which is NP-hard to manipulate without weights. 
Third, is there a connection
between the smoothness of the phase transition and
problem hardness? Sharp phase transitions
like that for 
satisfiability are associated with hard
decision problems, whilst
smooth transitions are
associated with easy instances of NP-hard
problems and with polynomial problems like 2-colorability. 
Fourth, these results demonstrate that
empirical studies improve our understanding
of manipulation. 
It would be interesting to consider
similar studies for related problems like
preference elicitation \cite{waaai07,waamas08,lprvwijcai07,prvwijcai07,prvwkr08}.

\bibliographystyle{named}

\end{document}